\documentclass[letterpaper, 10 pt, conference]{ieeeconf}  % Comment this line out if you need a4paper

\IEEEoverridecommandlockouts                              % This command is only needed if 
\usepackage{graphicx} % for pdf, bitmapped graphics files

\title{\LARGE \bf
	Offline and Online calibration of Mobile Robot and SLAM Device for Navigation
}

\author{Ryoichi Ishikawa$^{1}$, Takeshi Oishi$^{1}$ and Katsushi Ikeuchi$^{2}$% <-this % stops a space
	\thanks{$^{1}$ Ryoichi Ishikawa and Takeshi Oishi are with Institute of Industrial Science, The University of Tokyo, Japan
		{\tt\small \{ishikawa, oishi\}@cvl.iis.u-tokyo.ac.jp}}%
	\thanks{$^{2}$Katsushi Ikeuchi is with Microsoft, USA,
		{\tt\small katsuike@microsoft.com}}%
}
\begin{document}

\maketitle
\thispagestyle{empty}
\pagestyle{empty}

\begin{abstract}
	Robot navigation technology is required to accomplish difficult tasks in various environments. In navigation, it is necessary to know the information of the external environments and the state of the robot under the environment. On the other hand, various studies have been done on SLAM technology, which is also used for navigation, but also applied to devices for Mixed Reality and the like.
	
	In this paper, we propose a robot-device calibration method for navigation with a device using SLAM technology on a robot. The calibration is performed by using the position and orientation information given by the robot and the device. In the calibration, the most efficient way of movement is clarified according to the restriction of the robot movement. Furthermore, we also show a method to dynamically correct the position and orientation of the robot so that the information of the external environment and the shape information of the robot maintain consistency in order to reduce the dynamic error occurring during navigation.
	
	Our method can be easily used for various kinds of robots and localization with sufficient precision for navigation is possible with offline calibration and online position correction. In the experiments, we confirm the parameters obtained by two types of offline calibration according to the degree of freedom of robot movement and validate the effectiveness of online correction method by plotting localized position error during robot's intense movement. Finally, we show the demonstration of navigation using SLAM device.
	
\end{abstract}

\section{Introduction}

%Robot navigation spread
 Terrestrial mobile robots such as industrial AGV and home robot have been more popular and practical in various scenes. These robots are required to perform tasks autonomously on behalf of humans in flatten floor or ground. Robot navigation is a very important technique for the automatic execution of many tasks. Navigation requires information on the external environment and position and orientation of the robot in the environment. 

%Conventional technique
There is a method of localizing a robot on an environment map of a place in where the robot is navigated prepared in advance. Various algorithms are used to localize robot such as ICP algorithm \cite{lu1997robot}, 2D Monte Carlo Localization (MCL) \cite{fox1999monte}, 3D MCL, A vision-based approach \cite{ido2009indoor}, RGB-D camera and particle filter \cite{winterhalter2015accurate}. For estimating position and orientation, some kinds of landmarks such as RFID (Radio-frequency identifier) \cite{park2009autonomous} and two-dimensional bar code \cite{George2013} are also used.

% SLAM technique
Meanwhile, SLAM (Simultaneous Localization and Mapping) technology using various kinds of sensor such as monocular camera \cite{engel2014lsd,mur2015orb} and RGB-D camera \cite{endres20143} has been developed recently. SLAM-based navigation methods using various sensors such as laser range finder \cite{misono2007development,klanvcar2014mobile} and RBG-D camera \cite{oliver2012using,wang2016localization} are also proposed. SLAM is also applied to devices for Augmented Reality (AR) and Mixed Reality (MR) such as a head-mounted display. 

\begin{figure}[t]
	\begin{center}
		\includegraphics[width=\linewidth]{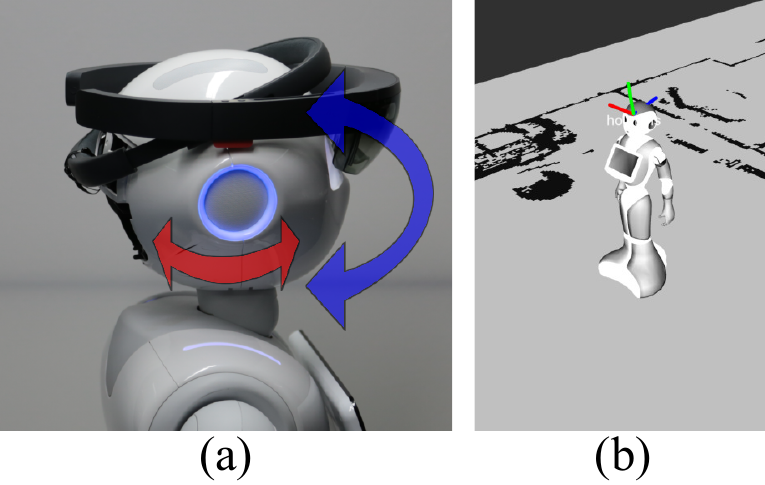}
		\caption{(a) Attaching SLAM device on robot and calibrating relative pose by moving robot head, (b)After calibration, SLAM device is located in front of robot head and the robot is localized in the map coordinate system.}
		\label{fig:introstep}
	\end{center}
\end{figure}

% SLAM device
We refer to a real-time three-dimensional sensing device as "SLAM device". The SLAM device has the functions of sensing, mapping external environment and estimating the self position and pose in real time. MR applications require this functions to display virtual objects in a fixed position even when the device moves. In this research, we deal with robot-SLAM device calibration to navigate robot by using SLAM function of SLAM device as shown in Fig.~\ref{fig:introstep}~(a).

% aim
The aim of our research is robot-sensor extrinsic calibration. This calibration is necessary because the robot itself knows only the internal state and have to know the position and orientation in the environment through the SLAM device. Calibration between camera and robot arm is well known as hand-eye calibration. In \cite{shiu1989calibration,park1994robot}, a mathematical solution of the equation $AX=XB$ is shown. Fassi, et al. similarly discuss the solution of $AX = XB$ from a geometrical point of view \cite{fassi2005hand}. Calibration methods of camera and IMU using the Kalman filter are also proposed in \cite{kelly2009fast, hol2010modeling}. However, these studies have not considered a restriction on freedom of the robot movements.

%where $X$ is the unknown $4\times4$ matrix including the rotation and translation between the part of the robot and the camera, and $A$ and $B$ are the known $4\times4$ matrices representing a rotation and translation of a transition when moving the robot and the camera respectively.

% Probrems
Some kinds of robot traveling ground have a restriction of movement. For example, a rover does not have a vertical rotation function. Completing calibration requires to know what parameters derived from a certain motion. Another problem is that dynamic errors may occur during navigation due to the shift of the device attached to the robot or the time lag of the encoders when joints of the robot moves. These errors can cause large localization error.

%Proposed method 1
In this paper, we clarify the necessary movement and information for the calibration and show the most efficient calibration method with respect to the degree of freedom of the robot movement. The aim of this system is to apply SLAM system to various types of robots including a robot having such a restriction on freedom of their movement that normal hand-eye calibration methods do not consider. We deal with two cases: a case where the robot can rotate in two directions, a horizontal direction and a vertical direction, and a case where the robot can rotate in the horizontal direction but cannot rotate in the vertical direction, and show the optimal calibration method for each case. In the latter case, a calibration parameter that cannot be obtained from position and pose transition is acquired by using the position information of the SLAM device as supplementary information.

%Proposed method 2
In addition to this offline calibration, we propose an online position correction method by adjusting the relative position and orientation of the robot and the device so that the consistency of the device-robot-external environment information holds. For example, the footprint pose of robot should be localized to be perpendicular to the floor. The rotation component of the calibration parameter is adjusted dynamically during robot moving to alleviate this localization error under the premise that the robot stands vertically and a vector perpendicular to the floor and a vector perpendicular to the ground plane of the robot coincide with each other.

%Our calibration method between SLAM device and robot can be easily applied to various kinds of robots even when the robot can not rotate in more than two directions. By reducing the error using online calibration parameter adjustment, sufficient accuracy for indoor navigation is also obtained.

%=====================================================================================================================
\section{Problem Setting and Notation}

The problem to solve in this calibration is to calculate relative rotation and translation parameters between "head" and "sd". To solve this problem, we use each self-position estimated by robot and SLAM device as inputs and position and orientation transitions of the SLAM device and robot are calculated respectively. The robot has the IMU and motor encoder etc. and it is assumed that the position and pose parameter of the robot is given. In the SLAM device, it is assumed that external information can be acquired and its own position and pose in the world coordinate system is given. Robot shape information and 3-D map of external environment obtained by SLAM device is also used for offline and online calibration.

In our system, several coordinate systems are used: "map" coordinate system, "sd" coordinate system of SLAM device, "head" coordinate system of the robot frame that SLAM device mounted on, and "foot" coordinate system of the part of the robot contacting the floor. For convenience, the coordinate system of the robot frame to which the SLAM device is attached is called "head", but even if there is no head mechanism like the rover, this method can be used.

\begin{figure}[t]
	\begin{center}
		\includegraphics[width=.75\linewidth]{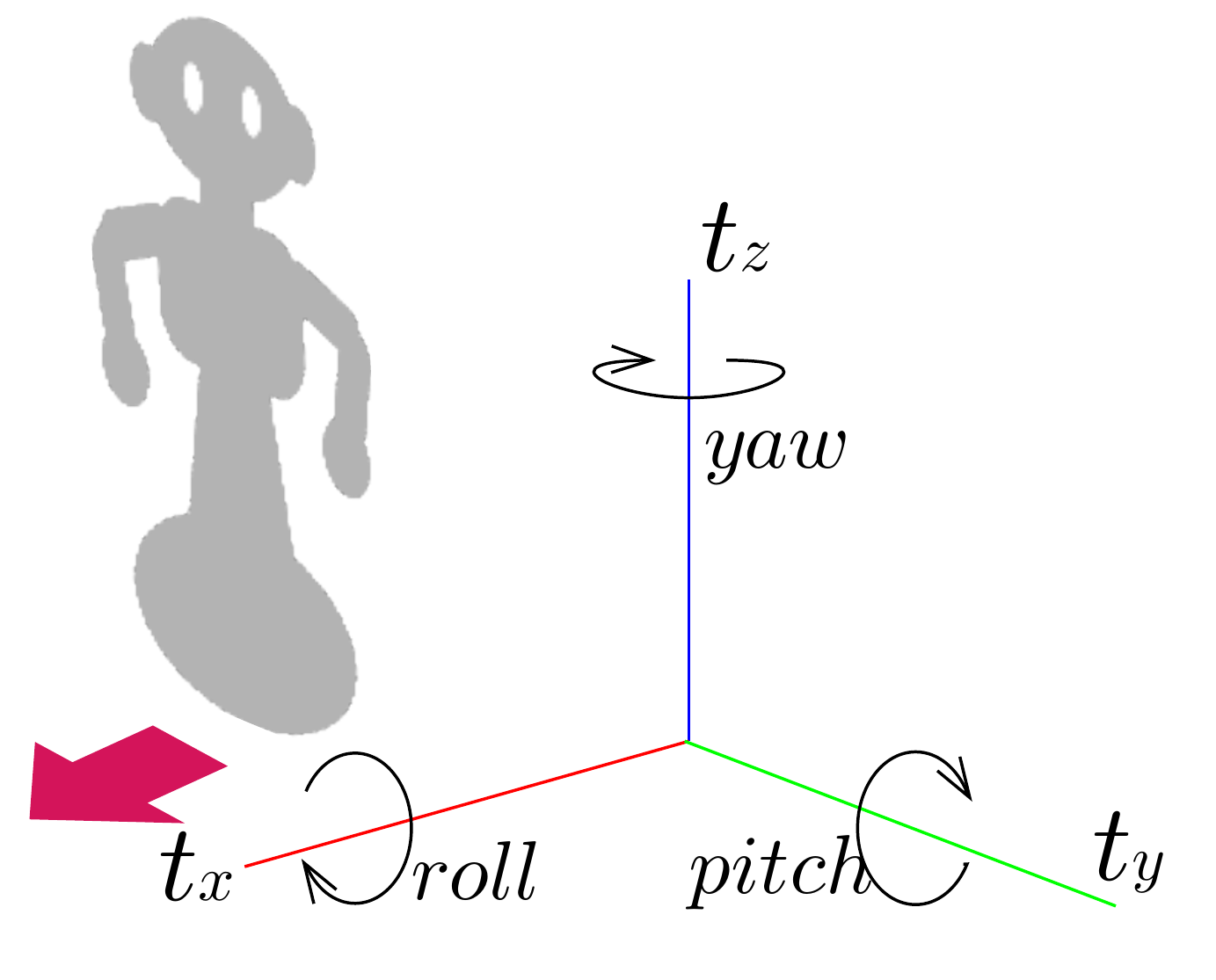}
		\caption{The relationship between the real world and the coordinate system. The forward direction of the robot is the x axis, and the vertical direction is the z axis.}
		\label{fig:coordinate}
	\end{center}
\end{figure}

We deal with the parameters of 6-DoF of an unknown rotational and translational parameter. As shown in Fig.~\ref{fig:coordinate}, we set the propulsion direction when the robot moves forward as $ x $ and the vertical direction of the robot as $ z $. $ t_x $, $t_y$, $ t_z $ is translation parameter along the axis $ x $, $ y $ , $ z $ and $ roll $, $ pitch $, $ yaw $ is rotation parameter around the axis $ x $, $ y $ , $ z $ respectively. 

%=========================================================================================================================================
\section{Analysis of Robot-Sensor Calibration from Transitions}
\label{seq:offline}
%Optimization calculation is performed by using the position and pose transition, and at that time,  The position information of the SLAM device is used when there is a limitation on the motion of the robot and the robot position and orientation transitions alone do not restrict all parameters. First, we discuss the relationship between position and pose transition and calibration parameter can be obtained by it. The position information of the SLAM device is used when there is a limitation on the motion of the robot and the robot position and orientation transitions alone do not restrict all parameters.

\begin{figure}[t]
	\begin{center}
		\includegraphics[width=.70\linewidth]{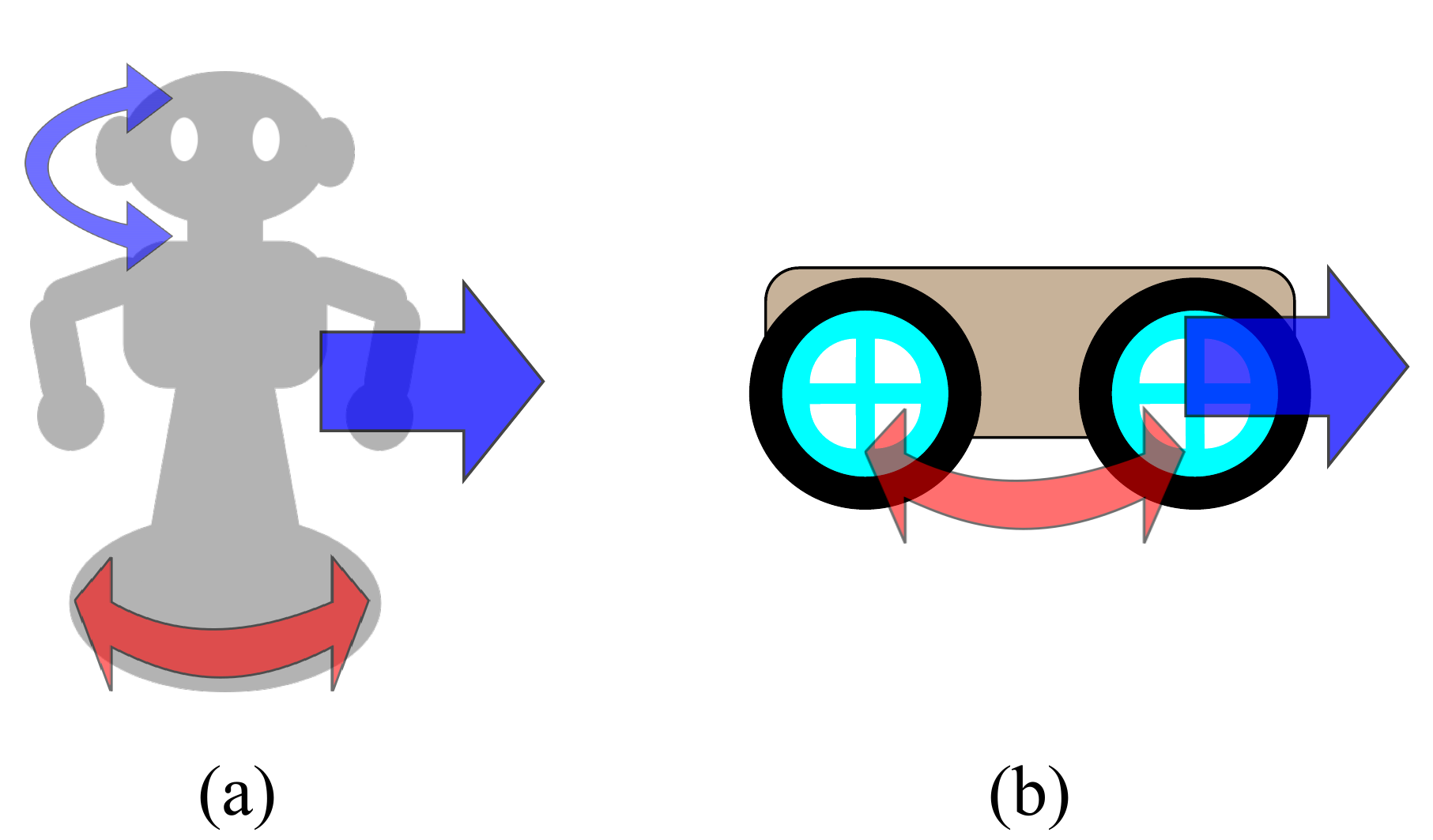}
		\caption{(a) An example of a robot capable of both moving and bi-directional rotation (b)An example of a robot that does not have a mechanism that rotates in the vertical direction}
		\label{fig:motlim}
	\end{center}
\end{figure}

We analyze what parameters can be obtained from a certain motion. In contrast to normal hand-eye calibration, there is a possibility that the robot's motion may be restricted depending on the robot to be used. For example, as shown in Fig.~\ref{fig:motlim}, a robot such as a humanoid type can swing the neck vertically and horizontally in addition to position movement. On the other hand, a robot like a rover type can not rotate in the longitudinal direction. We describe an optimal calibration method according to such degrees of freedom of movement.

\begin{figure}[t]
	\begin{center}
		\includegraphics[width=.70\linewidth]{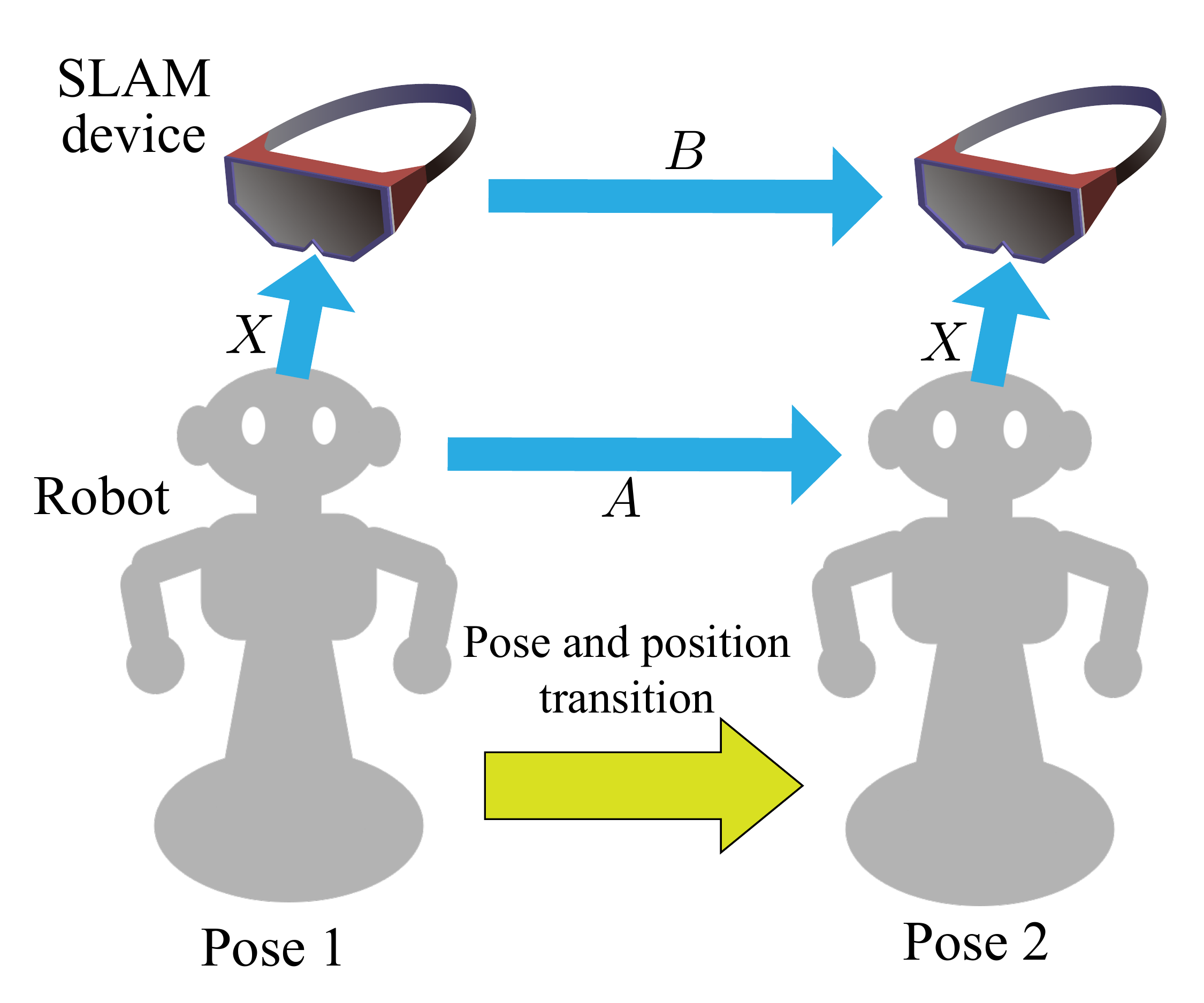}
		\caption{Schematic diagram of position and orientation transition.}
		\label{fig:overviewoff}
	\end{center}
\end{figure}

%=========================================================================================================================================
\subsection{Obtaining pose transition} 
% x,y,z--> tx, ty, tz

First, we explain the method to compute transition of robot and SLAM device for calculating the calibration parameters. In order to obtain the relative position and orientation between the robot and the SLAM device, the position and orientation transition is made and the differences between the two position and orientation parameters before and after the transition are taken. Let pose 1 be the initial posture of the head and the SLAM device and let pose 2 be their poses after the transition. Let ${\bf M_1}_{head}$ and ${\bf M_1}_{sd}$ be $4\times4$ matrix representing the position and pose in the local coordinates of each robot and SLAM device at pose 1 respectively. ${\bf M_2}_{head}$ and ${\bf M_2}_{sd}$ be $4\times4$ position and pose matrix in the local coordinates at pose 2 respectively. These matrices are the observed values. The differences between the two position and pose before and after the transition are given by the following equation,
\begin{eqnarray}
{\bf A}&=&{{\bf M_1}_{head}}^{-1} {\bf M_2}_{head}\\
{\bf B}&=&{{\bf M_1}_{sd}}^{-1} {\bf M_2}_{sd}
\end{eqnarray}
% add explanation of A, B, X, M is observed
${\bf A}$ indicates transformation of the head from pose 1 to pose 2, and ${\bf B}$ represents transformation of the SLAM device. Let ${\bf X}$ be an unknown $4\times4$ matrix representing the relative position and orientation between the robot and sensor given by the following equation,
\begin{eqnarray}
{\bf X}=\left[
\begin{array}{cc}
{\bf R} & {\bf t}\\
{\bf 0} & 1
\end{array}
\right].
\end{eqnarray}

As shown in Fig.~\ref{fig:overviewoff}, considering transformation from the coordinate frame of "head" at pose 1 to the coordinate frame of "sd" at pose 2, there are two transformations via "head" at pose 2 and via "sd" at pose 1. The former is represented by matrix calculation ${\bf AX}$ and the latter is represented by ${\bf XB}$. The results are same in both coordinate transformations, therefore ${\bf AX}={\bf XB}$ holds.

\subsection{Relation between movement and solved parameter}
%We consider the problem of calibration between SLAM device and robot into the problem of ${\bf AX}={\bf XB}$ and clarify robot transitions that can be used for good ${\bf X}$ estimation.
%For $ {\bf R} $, constraint is imposed on rotation parameters in two directions other than the direction of $ ({\bf I} - {\bf R}_A) {\bf t} $. However, the length of $ ({\bf I} - {\bf R}_A) {\bf t} $ depends on robot movement and how the SLAM device is attached to the robot, and if it is short, the constraint becomes weak and is not reliable for use in calibration.
%separate A, B
We describe the parameters obtained by the transitions. We mainly consider three types of movement of horizontal rotation, vertical rotation, forward movements and discuss what parameters can be derive from each motion based on solving ${\bf X}$ in ${\bf AX}={\bf XB}$. The solution of ${\bf AX}={\bf XB}$ is discussed in \cite{shiu1989calibration,park1994robot,fassi2005hand} and we can calculate ${\bf X}$ by using linear and non-linear optimization from ${\bf AX}={\bf XB}$. Let ${\bf R}_A$ and ${\bf R}_B$ be $3 \times 3$ rotation matrix components in ${\bf A}$ and ${\bf B}$ respectively, ${\bf k}_A$ and ${\bf k}_B$ be unit vectors indicating the rotation axis in ${\bf R}_A$ and ${\bf R}_B$ respectively. Let ${\bf t}_A$ and ${\bf t}_B$ be three-dimensional translation vector components. The following two formulas hold from ${\bf AX}={\bf XB}$ \cite{shiu1989calibration},
\begin{eqnarray}
\label{eq:axxbrot}
{\bf k}_A&=&{\bf R} {\bf k}_B,\\
\label{eq:axxbtran}
{\bf R}_A{\bf t}+{\bf t}_A&=&{\bf R}{\bf t}_{B}+{\bf t}.
\end{eqnarray}
First of all, consider the constraint obtained from Eq.~\ref{eq:axxbrot}. Given a set of ${\bf k}_A$, ${\bf k}_B$, the 2-DoF rotational parameters other than around the ${\bf k}_A$ vector among the 3-DoF rotation parameters included in ${\bf R}$ are determined. 

Subsequently, let us consider the constraints obtained from Eq.~\ref{eq:axxbtran}. Eq.~\ref{eq:axxbtran} can be transformed as follows,
\begin{eqnarray}
\label{eq:axxbtran_trans}
({\bf I}-{\bf R}_A){\bf t}={\bf t}_A-{\bf R}{\bf t}_{B}.
\end{eqnarray}
In Eq.~\ref{eq:axxbtran_trans}, rank of $({\bf I}-{\bf R}_A)$ is two. When rotating the robot and considering the case where ${\bf t}$ is decomposed into ${\bf k}_A$ components and two unit vectors ${\bf t}_1$ and ${\bf t}_2$, which is orthogonal to ${\bf k}_A$ and mutually orthogonal, ${\bf t}$ in Eq.~\ref{eq:axxbtran_trans} has a degree of freedom in the ${\bf k}_A$ direction and has constraints in the ${\bf t}_1$ and ${\bf t}_2$ directions since $({\bf I}-{\bf R}_A){\bf k}_A={\bf 0}$.

\subsubsection{Horizontal Rotation}
When the robot rotates horizontally, rotational axis directs in the $z$ direction. In the case of rotation, $ {\bf k}_A $ in Eq.~\ref{eq:axxbrot} directs the $ z $ axis, and we can obtain the parameters $ roll $, $ pitch $ other than the rotation $yaw$ around the $z$ axis. In the case of translation parameters, constraints based on Eq.~\ref{eq:axxbtran_trans} are mainly applied to $ t_x $ and $ t_y $.

\subsubsection{Vertical Rotation}
In the case of performing a rotational motion in the vertical direction (here, it is assumed to rotate around the $ y $ axis) such as swinging the neck vertically in the humanoid robot, similarly to horizontal rotation, two parameters of $ roll $ and $ yaw $ can be obtained from Eq.~\ref{eq:axxbrot}. Translation parameter $ t_x $ and $ t_z $ is also obtained from Eq.~\ref{eq:axxbtran_trans}.

\subsubsection{Forward Movement}
\begin{figure}[tb]
	\begin{center}
		\includegraphics[width=\linewidth]{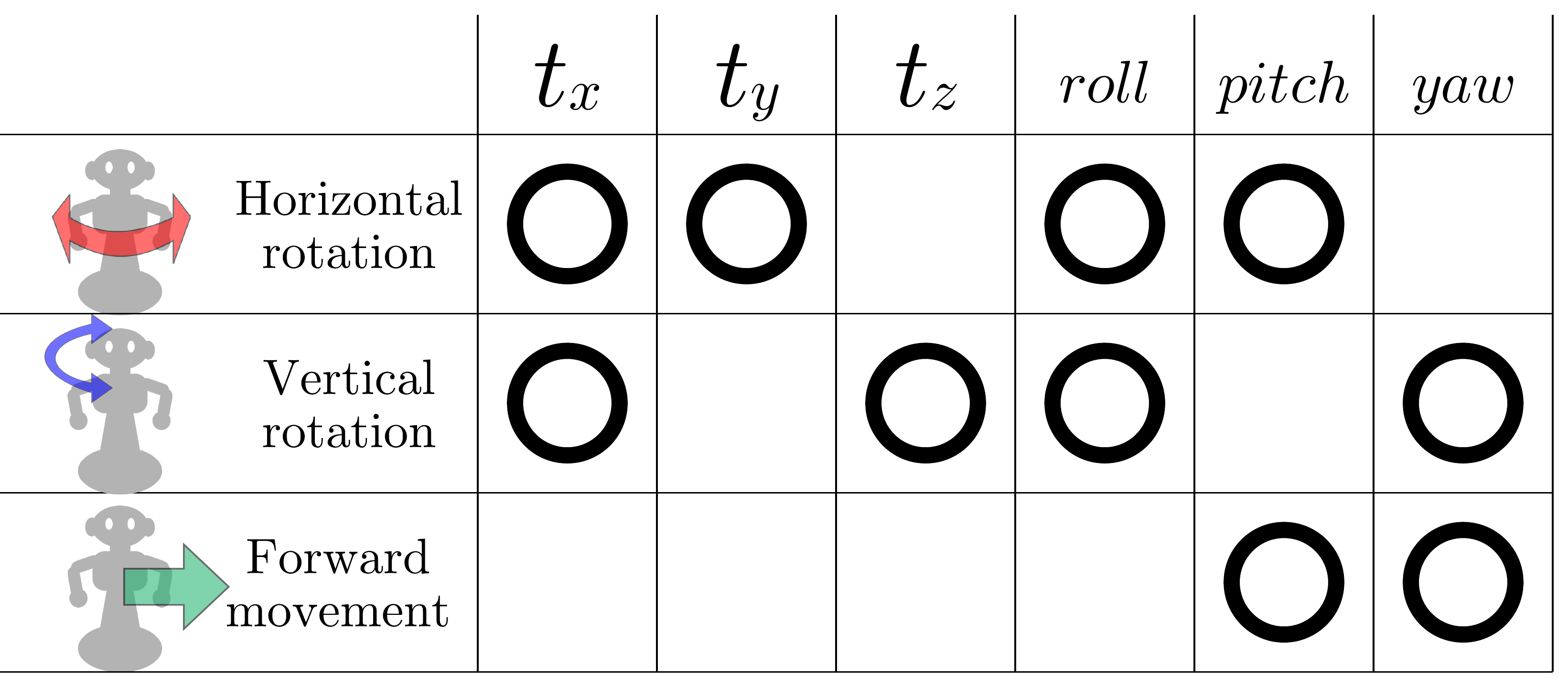}
		\caption{Relationship between movement and restricted parameter.}
		\label{fig:table}
	\end{center}
\end{figure}

When moving the robot straight ($ {\bf R}_A \approx {\bf I} $), Eq.~\ref {eq:axxbtran} becomes as follows,
\begin{eqnarray}
\label{eq:axxbtran_trans_app}
{\bf t}_A={\bf R}{\bf t}_{B},
\end{eqnarray}
This equation is the same form as Eq.~\ref{eq:axxbrot}. Therefore constraint is applied to rotation parameters $ pitch $, $ yaw $ when advancing the robot. Figure.~\ref{fig:table} shows these relations between the position and pose transition method (horizontal rotation, vertical rotation, forward) and the restricted parameters.

% As described above, both ${\bf R}$ and ${\bf t}$ can be obtained if there are two times position and orientation transitions with different rotation directions.

%Therefore, the constraint is imposed on two parameters by one equation. Translation components ${\bf t}_A$ and ${\bf t}_B$ of position and orientation transition constrain the rotation parameter ${\bf R}$ and rotational component ${\bf R}_A$ is necessary to obtain ${\bf t}$. 

%=========================================================================================================================================
\section{Off-line robot-SLAM device calibration}
We deal with two types of a robot: one is a robot that can rotate in two directions, the other is a robot that can move only in the horizontal direction ($x$ direction, $y$ direction, rotation around $yaw$ axis in Fig.~\ref{fig:coordinate}). The both are assumed to be capable of turn on the spot. In the former case, bi-directional rotation is optimal. In the later case, only five parameters can be obtained by horizontal rotation and forward movements. The remaining one is obtained using auxiliary information such as height of SLAM device. 

\subsection{Calibration using bi-directional rotation}
\label{seq:axxb}
We describe the conditions of position and orientation transition and the calibration method when the frame of a robot equipped with SLAM device can rotate in the vertical direction. We can see in Fig.~\ref{fig:table} that the constraint is applied to $ t_x $, $ t_y $, $ roll $, $ pitch $ with horizontal rotation. For the remaining $ t_z $ and $ yaw $, you can restrain these parameters by rotating in the vertical direction. Therefore, if horizontal rotation and vertical rotation are performed at least once, the constraint can be applied to all 6-DoF parameters.

\subsection{Calibration using horizontal movement and SLAM device height}
In the case of using a robot that is difficult to obtain $t_z$. For the component in the remaining $z$ direction, we cannot constrain it by horizontal position and pose transition alone, therefore we use the height of the SLAM device from the floor to calculate the $z$ parameter. Let $ {\bf n} $ be the normal vector of the floor and $ h $ be the height of the SLAM device from the floor. $ {\bf n} $, $ h $ can be observed from the environmental map of the SLAM device. Let ${\bf b}$ be a vector from "head" to "foot". It is known because robot's shape information is available. Let ${\bf o}$ be a parallel component to the floor of the vector from "sd" to "head". ${\bf o}$ indicates the shift of "sd" and "foot" in the plane parallel to floor.. It is perpendicular to ${\bf n}$ defined by the following equation,
\begin{eqnarray}
\label{eq:cntran1}
{\bf o}=h{\bf n}-{\bf R}^{-1}{\bf t}+{\bf b}.
\end{eqnarray}

From Eq.~\ref{eq:cntran1}, the following equation holds,
\begin{eqnarray}
\label{eq:cntran2}
{\bf n}\cdot{\bf o}={\bf n}\cdot(h{\bf n}-{\bf R}^{-1}{\bf t}+{\bf b})=0
\end{eqnarray}
This Eq.~\ref{eq:cntran2} can constrain the remaining $t_z$ in the vertical direction of the floor.

\subsection{The case of a complex movement}
Here, we consider a case where a complex position and pose transition including both rotation and translation at once is performed. From Eq.~\ref{eq:axxbtran_trans}, ${\bf t}$ is obtained using the difference between the translation component ${\bf t}_A$ of the robot's head and translational movement component ${\bf R}{\bf t}_{B}$ of the SLAM device. Now, when rotating and translating at the same time in position and pose transition at once, accumulation error between odometry of the SLAM device and the robot occurs. This accumulation error is directly related to the error of ${\bf t}$ calculated based on the difference between ${\bf t}_A$ and ${\bf R}{\bf t}_{B}$. Therefore, in order to prevent this accumulation error and perform accurate calibration, it is necessary to minimize the robot translational movement (= length of ${\bf t}_A$).

We consider whether there is superiority in terms of the minimum number of transitions necessary for calibration when performing position and pose transition including both rotation and translation at the same time. In Eq.~\ref{eq:axxbtran_trans}, although constraints are applied to the three parameters $t_x$, $t_y$, $yaw$, the rank of $({\bf I}-{\bf R}_A)$ is two and one equation can bind only two parameters of the three. Therefore, even if a complex position and pose transition is performed, at least two transitions are required, as in the case of when using movements with only rotation and with only translation. For these reasons, it is better to use the only rotation movement and only translation movement so that the influence of accumulation error can be avoided.

\section{On-line position adjustment}
\begin{figure}[t]
	\begin{center}
		\includegraphics[width=\linewidth]{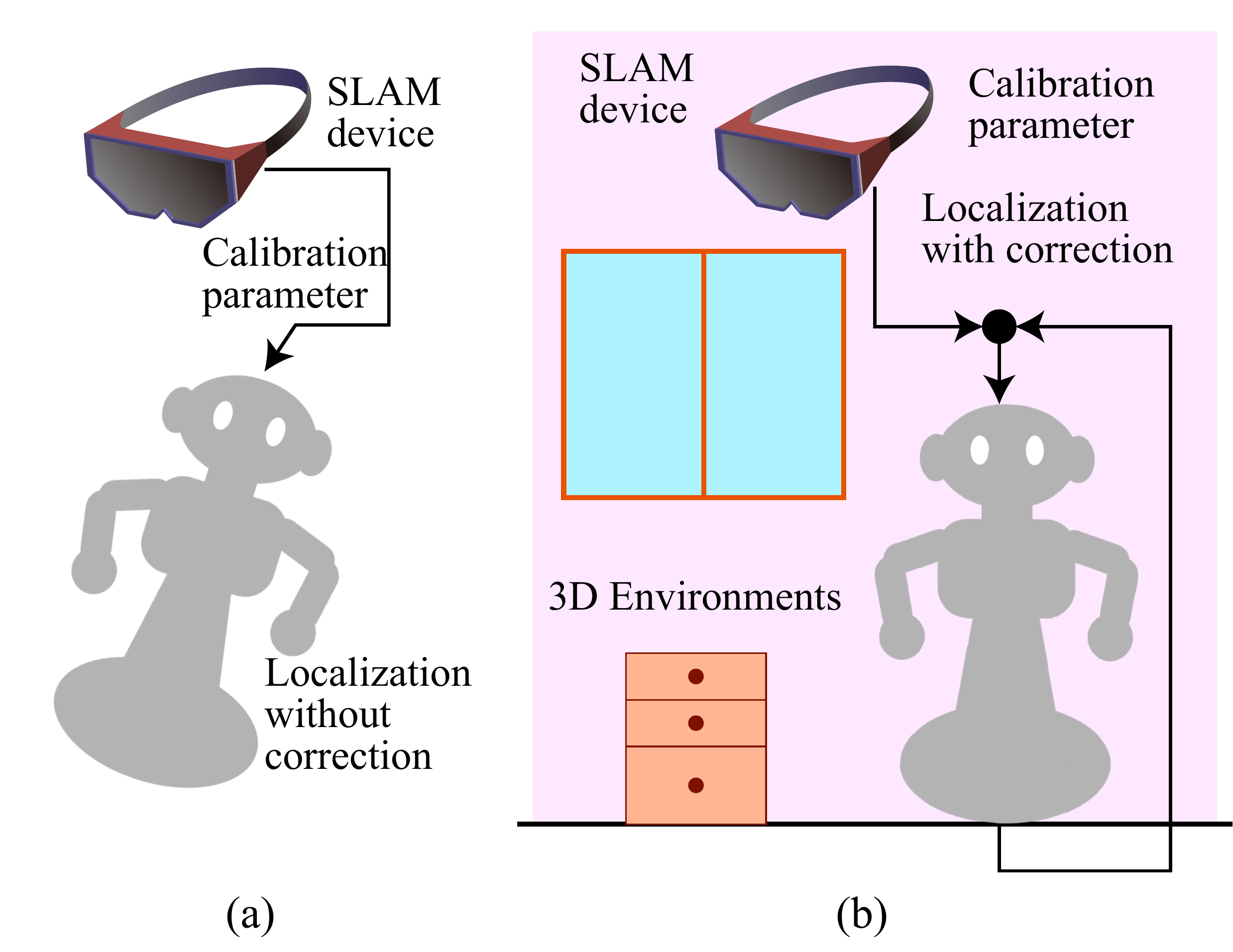}
		\caption{Overview of online adjustment. (a) When the error exists within transformation from SLAM device to robot's foot, the robot is localized diagonally. (b) Rotation error is corrected by making loop closure between SLAM device - robot shape - environment. Consistency is taken by aligning the perpendicular vector of the robot with floor normal vector.}
		\label{fig:overviewon}
	\end{center}
\end{figure}
\begin{figure}[t]
	\begin{center}
		\includegraphics[width=.75\linewidth]{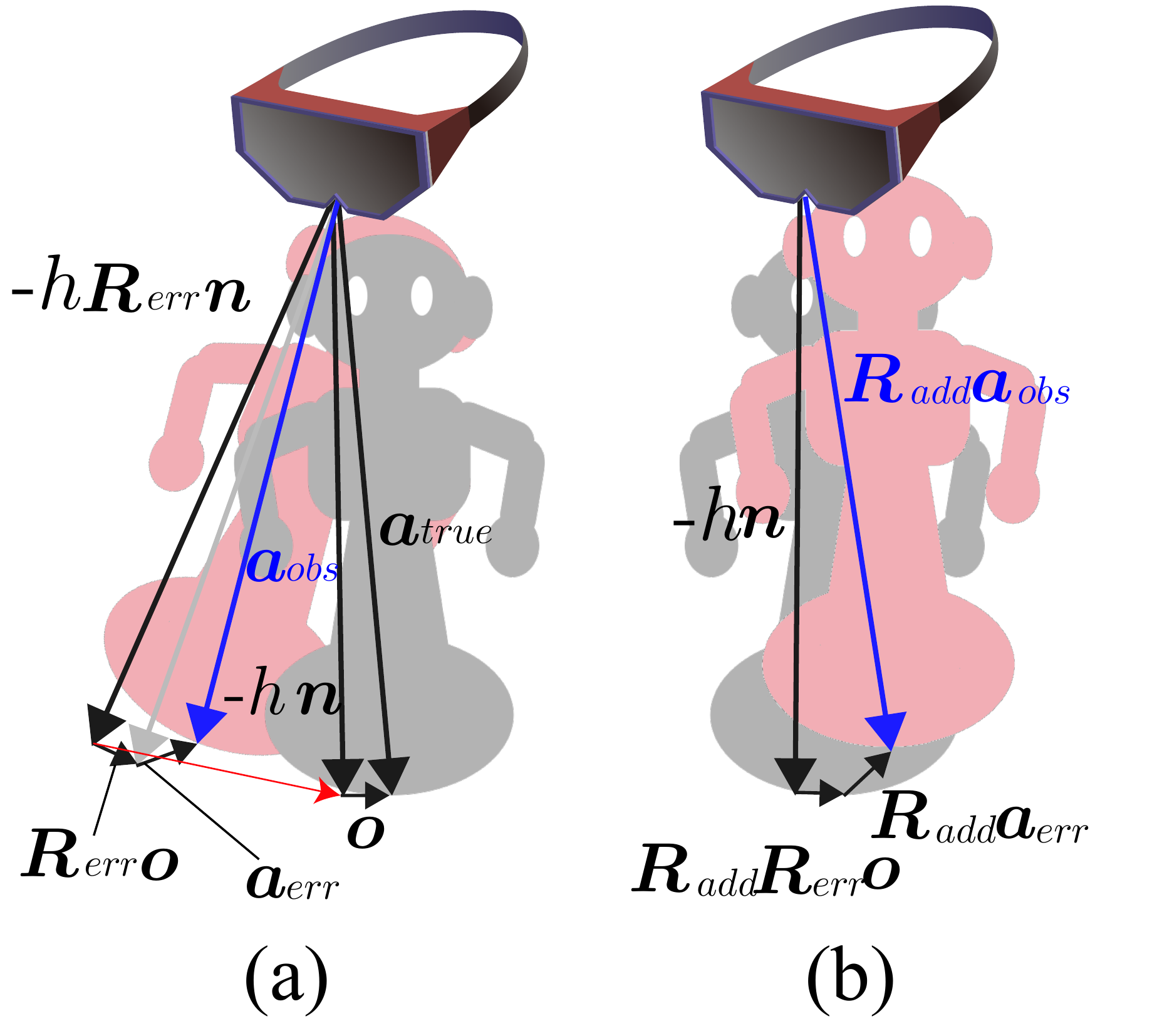}
		\caption{Schematic Diagram of position of the robot and SLAM device. The gray robot is in the correct positional relationship with the device and the red is the localized robot by SLAM device and calibration parameter. (a) The red robot is localized with error and the difference between $h{\bf R}_{err}{\bf n}$ and $h{\bf n}$ becomes the large part of the error of localization. (b) The position and pose of localized red robot is corrected by additional rotation and the difference between $h{\bf R}_{err}{\bf n}$ and $h{\bf n}$ is canceled.}
		\label{fig:calib}
	\end{center}
\end{figure}
We describe an online position correction method to correct relative position between the robot and the SLAM device so that the information of the external environment, the shape of the robot and the position of the device are consistent. Even if the calibration is performed offline in advance, there are several dynamic factors that make the location of the localized base incorrect during navigation. There is a possibility that the SLAM device may be misplaced while navigation is in operation. 
The accuracy of the IMU or the encoder of the robot causes an error, or the time lag of the encoder when the joint is moved also causes a large error. This makes robot be localized diagonally as shown in Fig.~\ref{fig:overviewon}~(a). Therefore we adjust the robot's pose so that the robot is properly grounded as shown in Fig.~\ref{fig:overviewon}~(b).

Among the errors caused by these causes, in particular, errors in the $roll$ axis and $pitch$ axis rotation lead to a large localization error when the device is mounted at a high position. In the online correction method, the parameters between the SLAM device and the robot are modified based on the premise that the robot stands perpendicular to the ground. The overview is shown in the Fig.~\ref{fig:overviewon}. As a specific calculation, first, from the information of the external environment acquired by the SLAM device, find the normal vector of the floor and the vector perpendicular to the ground plane of the robot localized by the SLAM device before correction. Then apply additional correction rotation to the coordinate transformation between the robot and the device so that the two vectors match. This operation creates a partial loop closure that constrains two parameters and alleviates the localization error.

\subsection{Error analysis}
We explain what factor become large localized error. A schematic diagram is shown in Fig.~\ref{fig:calib}~(a). Let $ {\bf a}_{true} $ be the true value of the vector from the "sd" coordinate center to the "foot" coordinate center and $ {\bf a}_{obs} $ be the observation value including translation and rotation error. Let ${\bf a}_{err}$ and ${\bf R}_{err}$ be the translational and rotational error factor. This error factor includes all the error in the transformations from "sd" to "foot" such as calibration error or robots encoder error. Using the translational error $ {\bf a}_{err} $ and the rotational error factor $ {\bf R}_{err} $, $ {\bf a}_{obs} $ is expressed as follows,
\begin{eqnarray}
\label{eq:t_obs}
{\bf a}_{obs}={\bf R}_{err}{\bf a}_{true}+{\bf a}_{err}.
\end{eqnarray}
It is assumed that this error factor is caused by the above-mentioned deviation of the SLAM device and the error of the encoder. Next, decompose ${\bf a}_{true}$ into a component in the direction of the unit vector ${\bf n}$ perpendicular to the floor and a vector ${\bf o}$ parallel to the floor and orthogonal to ${\bf n}$,
\begin{eqnarray}
\label{eq:t_dec}
{\bf a}_{true}=-h{\bf n}+{\bf o},
\end{eqnarray}
where,
\begin{eqnarray}
\label{eq:decomp}
h=-{\bf a}_{true}\cdot{\bf n}.
\end{eqnarray}

Substitute Eq.~\ref{eq:t_dec} for Eq.~\ref{eq:t_obs},
\begin{eqnarray}
\label{eq:t_obs2}
{\bf a}_{obs}=-h{\bf R}_{err}{\bf n}+{\bf R}_{err}{\bf o}+{\bf a}_{err}.
\end{eqnarray}
When comparing Eq.~\ref{eq:t_dec} and Eq.~\ref{eq:t_obs2}, the difference between $h{\bf R}_{err}{\bf n}$ and $h{\bf n}$ (The red line in Fig.~\ref{fig:calib}~(a)) becomes the large part of the error of localization in the case of a particularly tall robot because localized error is proportional to the sine of the rotation error and the height of the robot.

\subsection{Error correction}
We explain how the error is reduced when applying an additional correction rotation. A schematic diagram is shown in Fig.~\ref{fig:calib}~(b). Letting a vector perpendicular to the ground plane of the localized robot before correction be ${\bf n'}$,
\begin{eqnarray}
\label{eq:nerr}
{\bf n'}={\bf R}_{err}{\bf n}.
\end{eqnarray}
Here the rotation matrix ${\bf R}_{add}$ to rotate ${\bf n'}$ in the same direction as ${\bf n}$ is decided, ie ${\bf R}_{add}$ satisfies the following equation:
\begin{eqnarray}
\label{eq:ncor}
{\bf n}={\bf R}_{add}{\bf n'}.
\end{eqnarray}
Integrating this correction matrix into the observation value ${\bf a}_{obs}$,
\begin{eqnarray}
\label{eq:t_cor}
{\bf R}_{add}{\bf a}_{obs}&=&-h{\bf R}_{add}{\bf R}_{err}{\bf n}+{\bf R}_{add}({\bf R}_{err}{\bf o}+{\bf a}_{err})\nonumber\\
&=&-h{\bf n}+{\bf R}_{add}({\bf R}_{err}{\bf o}+{\bf a}_{err}),
\end{eqnarray}
where, we used that the following equation is established from Eq.~\ref{eq:nerr} and Eq.~\ref{eq:ncor}
\begin{eqnarray}
\label{eq:nnochange}
{\bf n}={\bf R}_{add}{\bf R}_{err}{\bf n},
\end{eqnarray}

Comparing Eq.~\ref{eq:t_dec}, Eq.~\ref{eq:t_obs2}, Eq.~\ref{eq:t_cor}, we can see that the term of $h{\bf R}_{err}{\bf n}$ in Eq.~\ref{eq:t_obs2}, which was the cause of the error increase, was corrected and canceled in Eq.~\ref{eq:t_dec} and Eq.~\ref{eq:t_cor}. Schematic diagram of this is shown in Fig.~\ref{fig:calib}. This correction method cancels $roll$ and $pitch$, witch are the two degrees of freedom rotational parameters other than the rotation component around the vector ${\bf n}$ of ${\bf R}_{err}$ and removes the error component of localization proportional to the height of the robot.

When actually calculating ${\bf R}_{add}$, we can use the vector perpendicular to the ground plane of the localized robot before correction as the observed value of $n'$ and the floor normal obtained from the SLAM device's 3D environmental map as the observed value of $n$. Although ${\bf R}_{add}$ is not unique, the rotation matrix with ${\bf n}\times{\bf n'}$ as its rotation axis and the angle between ${\bf n}$ and ${\bf n'}$ as the rotation angle satisfies the condition as ${\bf R}_{add}$.

%=========================================================================================================================================

\section{Experimental results}

In the method evaluation, first, we check the values of parameters obtained in the two ways explained in Sec.~\ref{seq:offline}, then validate the online position correction method by confirming the error of the localized position when the joint of the robot is strongly moved. Finally, we demonstrate a robot navigation system in an indoor scene using SLAM devices.

\subsection{Implementation}
For experimental evaluation, we used SoftBank Pepper\footnote{https://www.softbank.jp/en/robot/} as the robot and Microsoft HoloLens\footnote{https://www.microsoft.com/hololens} as the SLAM device. Robot Operating System (ROS)\footnote{http://www.ros.org/} is used as the host system. HoloLens is attached to the head of Pepper as shown in Fig.~\ref{fig:introstep}~(a). HoloLens has a function to create a 3D map of the external environment and to record any point in real space as a "spatial anchor" with orientation in device memory. In this time, To make it easy to incorporate obstacles and other information into the system, navigation is done using a pre-made 2D map.

Regarding implementation, in order to localize HoloLens in ROS's world coordinate system, we align the image obtained by rendering a three-dimensional map inside HoloLens from an orthogonal viewpoint from the top with the pre-made 2D floor map. From the result of this alignment, the spatial anchor can be placed in the coordinate system of ROS and the HoloLens can be localized on the map. Even if HoloLens is localized once by setting up a spatial anchor, there is a tendency that the deviation between the position of HoloLens in the real world and the localized HoloLens in 2D map becomes large due to the error between the environment map inside HoloLens and the floor map. In order to avoid this error, When HoloLens moves away from the spacial anchor which is currently the standard, alignment of the 3D map inside HoloLens to the floor map with the current HoloLens position as the initial position is performed again to eliminate the accumulation error and a new spatial anchor is localized in the floor map.

\subsection{Off-line Calibration}

\begin{table}[htb]
	\label{tb:noise}
	\begin{center}
		\caption{Noise and bias values in simulation environments}
		\begin{tabular}{|c|c|}
			\hline
			Joint angle&$0.001(rad)$\\
			\hline
			SLAM Device position&$0.002(m)$\\
			\hline
			SLAM Device orientation&$0.004(rad)$\\
			\hline
			Floor point& $0.02(m)$\\
			\hline
			$\gamma_1$ & $0.04$\\
			\hline
			$\gamma_2$ & $0.04$\\
			\hline
			$a$ & $0.985$\\
			\hline
			$b$ & $0.01$\\
			\hline
		\end{tabular}
	\end{center}
\end{table}

\begin{table*}[htb]
	\begin{center}
		\caption{Calibration parameter value obtained in simulation environments}
		\begin{tabular}{|c|c||c|c|c|}
			\hline
			Calibration method&&pos error(m)&$x$ axis angle error(rad)&$y$ axis angle error(rad)\\
			\hline
			&average&0.009589&0.013243&0.011954 \\
			\cline{2-5}
			two-way rotation&median&0.011113&0.010341&0.009204 \\
			\cline{1-5}
			&average&0.006221&0.007964&0.006927 \\
			\cline{2-5}
			horizontal movement&median&0.005444&0.007186&0.005484 \\
			\hline
		\end{tabular}
	\end{center}
	\label{tb:sim}
\end{table*}

\begin{table*}[htb]
	
	\caption{Calibration parameter value obtained in real environments}
	\begin{center}
		
		\begin{tabular}{|c|c||c|c|c||c|c|c|c|}
			\hline
			&& x (m)& y (m)& z (m)& angle(rad) & axis x & axis y&axis z\\
			\hline
			two-way rotation&mean&0.083285&0.031271&0.129084&1.66594&0.511814&-0.49433&-0.70262\\
			\cline{2-9}
			
			&standard deviation&0.000531&0.000906&0.00169&0.000391&0.000877&0.000786&0.000232\\	
			\hline
			\hline
			horizontal movement&mean&0.101651&0.031098&0.11713&1.642449&0.52515&-0.47133&-0.70838\\
			\cline{2-9}
			&standard deviation&0.004845&0.002489&0.002671&0.010782&0.015054&0.00668&0.007034\\
			\hline
		\end{tabular}
	\end{center}
	\label{tb:calibpara}
\end{table*}

\subsubsection{Simulation environments}
First, we show the results of calibration using simulation environment. The simulation robot consists of a body and a head, and the head has two joints around $pitch$ axis and $yaw$ axis. The SLAM device is fixed with respect to the head. For the value of the joint angle issued by the robot, the position and posture given by the SLAM device, and the position of the floor (the coordinates of the foot of the perpendicular drawn from the SLAM device towards the floor), a model in which Gaussian noise is added to the true value is defined. For the odometry of the robot, we create a model considering the following two errors: the first one is an error between the speed value observed by a robot and velocity command value and the other is the difference between the velocity observed by the robot and the actual distance the robot has advanced. Let $dv(t)$ be the velocity value in the two-dimensional translation ($ dx, dy $) or angle ($ d\theta $) direction at time $t$, $ a_v $ be the acceleration or angular acceleration calculated from the commanded velocity value and the current observation speed value, $ G(a, b)$ be Gaussian disturbance with median $a$ and deviation $b$, $\gamma $ be a coefficient value, $ dv'(t +\Delta t) $ be the value of the displacement that actually changed during the time $t$ to $t+\Delta t$. The speed update formula and the displacement are defined as follows,
\begin{eqnarray}
\label{eq:sim_vel}
dv(t+\Delta t)&=&dv(t)+a_v \Delta t\nonumber\\
&+&G(0,\gamma_1|dv(t)|)+G(0,\gamma_2|dv(t)|)\\
\label{eq:sim_true_d}
dv'(t+\Delta t)&=&dv(t+\Delta t)G(a,b)\Delta t.
\end{eqnarray}

Table~\ref{tb:noise} show the set noise and bias values. The simulation update frequency is set to $100 Hz$. The height of the robot head joint from the ground is $ 1.1\: m $. In the position and orientation between the robot head and the SLAM device, the true value of the rotation is the identity matrix and the true value of translation is $ (0.12,\: 0.12,\: 0.12 )(m)$. In bidirectional calibration, we make two horizontal and two vertical transitions of rotating neck joints $ 0.3 rad $. In the calibration using the horizontal motion, two transitions to rotate on site for 2 seconds at $ 0.3 rad/sec $ and transitions to go straight for 2 seconds at $ 0.3 m / sec $ are performed.

Table~\ref{tb:sim} shows the result. Each calibration is performed five times and the average value and the median value of the error from the true value are recorded. In the case where the median error of translation is the largest, the median value of translation error is $ 1.11 cm $. This shows that if the dynamic error is ignored, it is sufficient to consider the centimeters order margin against the size of the robot. It can be said that this error is enough small for navigation of indoor scene.

\subsubsection{Real environments}
Next, we compare the values of parameters obtained by the calibration with two-way rotation transitions and the calibration with the horizontal position and pose transitions + HoloLens position information. In calibration using bidirectional rotation, we control the $ pitch $ angle and $ yaw $ angle of the neck and perform horizontal rotation transformation twice and vertical rotation transition twice in one calibration. In the case of using horizontal position and pose transitions, we manipulate Pepper's wheel and make twice rotational transitions and make two forward transitions while recording the height of HoloLens and the normal vector of the floor at five locations. Table.~\ref{tb:calibpara} shows the average value and standard deviation of parameters when each calibration method is performed five times. 

For the calibration using two-way rotation transitions, the standard deviation of translational movement is less than $ 2 mm $, and the angle is less than $ 4.0 \times 10^{-4} rad $. For the calibration using horizontal position and pose transition, the standard deviation of it is larger than the value in the case of two-way rotation calibration. However, it can be said that the accuracy is sufficient for navigation.

\subsection{On-line Adjustment}
\begin{figure}
	\begin{center}
		\includegraphics[width=.9\linewidth]{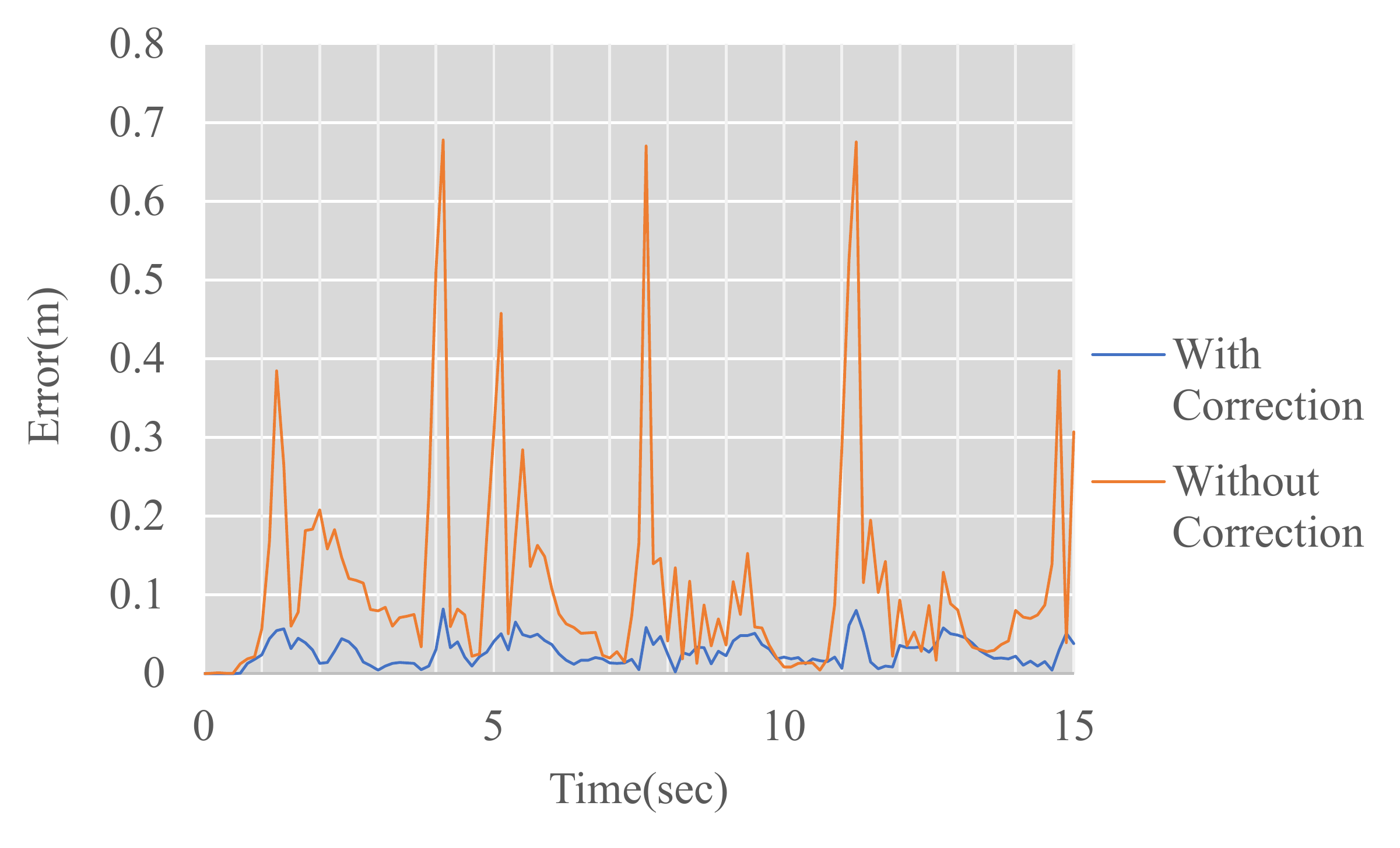}
		\caption{Localized position error during motion shaking Pepper head wearing HoloLens drastically. During the motion, pepper footprint do not move. Therefore ideally localized pepper footprint position is constant, however robot joint encoder error or time lug during motion causes to make localized footprint position unstable. During the movement, adjusted position has less than $10 cm$ error (Blue line) whereas there are more than $60 cm$ error in the result of before adjusted position (Orange line).  }
		\label{fig:onlineadjuster}
	\end{center}
\end{figure}

\begin{figure*}[t]
	\begin{center}
		\includegraphics[width=\linewidth]{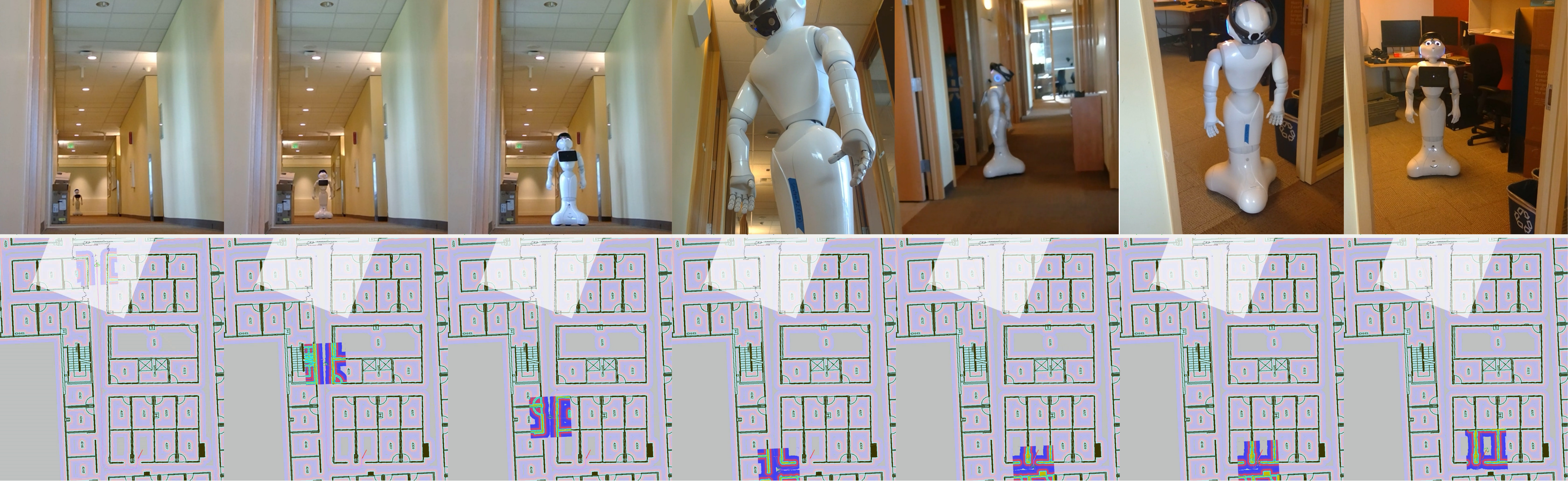}
		\caption{Navigation with our system using external SLAM device. Lower sequence is floor map in GUI. First, Pepper walks long corridor and then turn at the corner. Finally, Pepper enters a room and reaches to the destination.}
		\label{fig:navsequence}
	\end{center}
\end{figure*}
Next, we demonstrate the effectiveness of the online position correction method. We record both localized positions with and without correction in two dimensions during moving robot joints vigorously to validate our online correction method. While the robot is moving, the true value of the plotted foot position does not change in two dimensions since the joint moves but the position of the robot's foot has not changed. However, when the joint of the robot is violently moved, an encoder error due to a time lag occurs, which causes an error in observed localized positions via HoloLens.

Figure~\ref{fig:onlineadjuster} shows the result of plotting the error of the localized position of pepper's foot from the initial position during the operation of the joint of the robot. In the offline calibration, we use the two-way rotation calibration method and rotate in the horizontal direction and the vertical direction twice respectively. During operation, an error of at least $60 cm$ at maximum is recorded without correction, whereas the plot at the corrected position is less than $10 cm$. This result shows that the proposed position correction method can absorb the localization error.

\subsection{Navigation}
\begin{figure}
	\begin{center}
		\includegraphics[width=.8\linewidth]{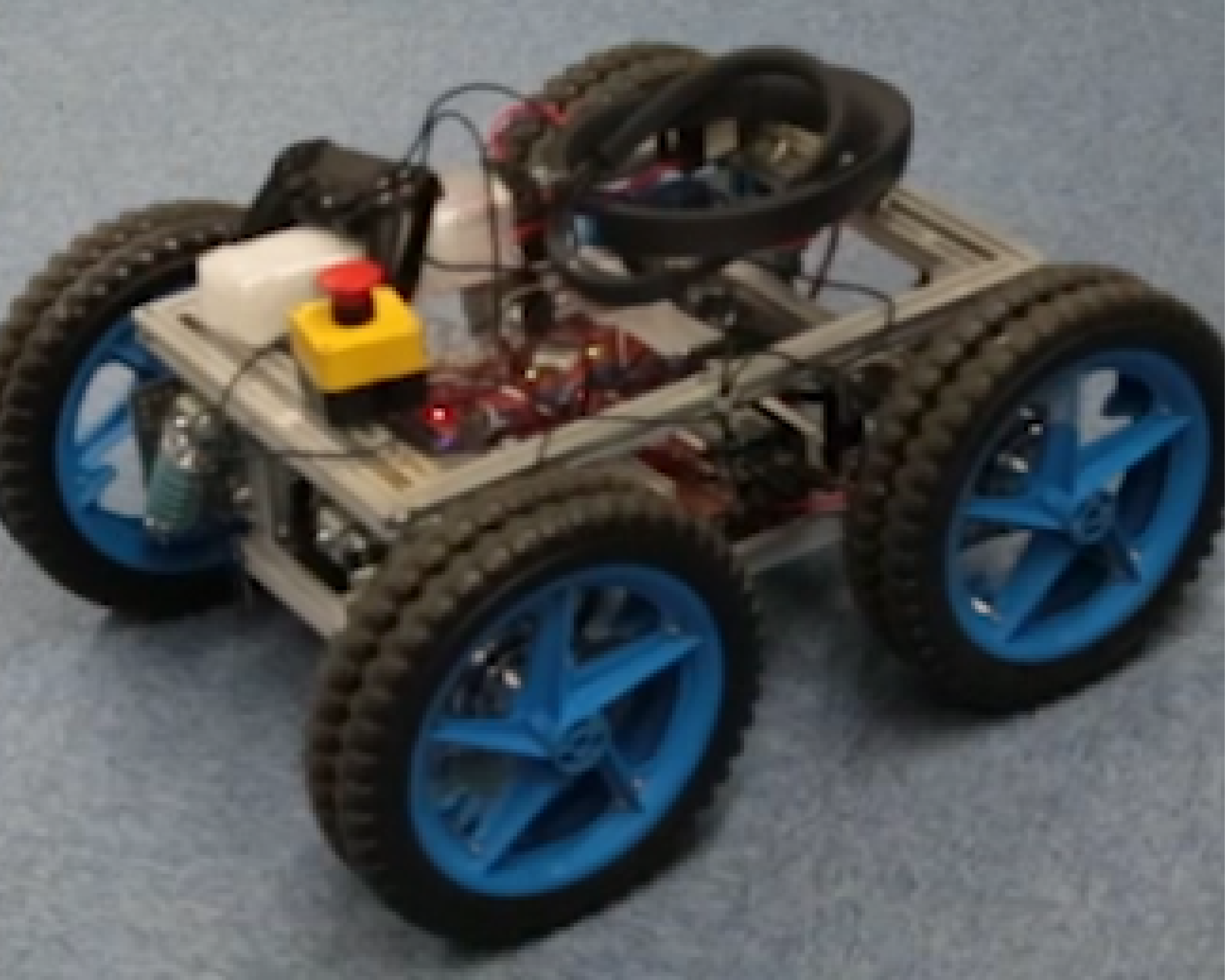}
		\caption{Rover Navigation using SLAM device}
		\label{fig:rovnav}
	\end{center}
\end{figure}
The steps for demonstrating the navigation system using Pepper and HoloLens are as follows: First, attach HoloLens to the head of Pepper, then establish communication with the host system and Pepper and HoloLens, and run navigation and localization program. After that, calibrate the Pepper's head and HoloLens using two-way rotation transitions (Fig.~\ref{fig:introstep}~(a) and (b)). Regarding the localization of HoloLens, by specifying the position of HoloLens on a two-dimensional floor map manually through a GUI, alignment is performed using it as an initial position and a spacial anchor is installed. Finally, navigation is executed by specifying the destination on the 2D map. As shown in Fig.~\ref{fig:rovnav}, it is also possible to navigate the rover by placing the SLAM device on it and applying calibration using horizontal movement

In the navigation, we make a global route plan using the Dijkstra method implemented in ROS's NavFn \footnote{http://wiki.ros.org/navfn} package, create a cost map \cite{lu2014layered}. We used Dynamic Window Approach \cite{fox1997dynamic} for local route planning. Figure.~\ref{fig:navsequence} shows a continuous photograph of navigation where Pepper goes through a long corridor and enters the room.

\section{Conclusion}
In this paper, we proposed a calibration method between external SLAM device and a robot for navigation. Although a humanoid type robot attached with a SLAM device is used for the demonstration, the proposed method can be easily applied to various robots having a self-position estimation function. In the experiments, 2D floor maps created in advance are used, however, this system can be extended to mapless navigation and three-dimensional navigation.

\section*{ACKNOWLEDGMENT}
Some part of the experiments was conducted with the assistance of the Strategic prototyping group, Microsoft. In particular, Yutaka Suzue's help was greatly appreciated. We also appreciate Yoshihiro Sato, Computer Vision Laboratory, The University of Tokyo. This work was partly supported by the social corporate program (Base Technologies for Future Robots) sponsored by NIDEC corporation and supported by JSPS Research Fellow Grant No.16J09277.

{\small
	\bibliographystyle{IEEEtran}
	\bibliography{myrefs}

\begin{thebibliography}{10}
\providecommand{\url}[1]{#1}
\csname url@samestyle\endcsname
\providecommand{\newblock}{\relax}
\providecommand{\bibinfo}[2]{#2}
\providecommand{\BIBentrySTDinterwordspacing}{\spaceskip=0pt\relax}
\providecommand{\BIBentryALTinterwordstretchfactor}{4}
\providecommand{\BIBentryALTinterwordspacing}{\spaceskip=\fontdimen2\font plus
\BIBentryALTinterwordstretchfactor\fontdimen3\font minus
  \fontdimen4\font\relax}
\providecommand{\BIBforeignlanguage}[2]{{%
\expandafter\ifx\csname l@#1\endcsname\relax
\typeout{** WARNING: IEEEtran.bst: No hyphenation pattern has been}%
\typeout{** loaded for the language `#1'. Using the pattern for}%
\typeout{** the default language instead.}%
\else
\language=\csname l@#1\endcsname
\fi
#2}}
\providecommand{\BIBdecl}{\relax}
\BIBdecl

\bibitem{lu1997robot}
F.~Lu and E.~Milios, ``Robot pose estimation in unknown environments by
  matching 2d range scans,'' \emph{Journal of Intelligent and Robotic systems},
  vol.~18, no.~3, pp. 249--275, 1997.

\bibitem{fox1999monte}
D.~Fox, W.~Burgard, F.~Dellaert, and S.~Thrun, ``Monte carlo localization:
  Efficient position estimation for mobile robots,'' \emph{AAAI/IAAI}, vol.
  1999, no. 343-349, pp. 2--2, 1999.

\bibitem{ido2009indoor}
J.~Ido, Y.~Shimizu, Y.~Matsumoto, and T.~Ogasawara, ``Indoor navigation for a
  humanoid robot using a view sequence,'' \emph{The International Journal of
  Robotics Research}, vol.~28, no.~2, pp. 315--325, 2009.

\bibitem{winterhalter2015accurate}
W.~Winterhalter, F.~Fleckenstein, B.~Steder, L.~Spinello, and W.~Burgard,
  ``{Accurate indoor localization for RGB-D smartphones and tablets given 2D
  floor plans},'' in \emph{Intelligent Robots and Systems (IROS), 2015 IEEE/RSJ
  International Conference on}.\hskip 1em plus 0.5em minus 0.4em\relax IEEE,
  2015, pp. 3138--3143.

\bibitem{park2009autonomous}
S.~Park and S.~Hashimoto, ``{Autonomous mobile robot navigation using passive
  RFID in indoor environment},'' \emph{IEEE Transactions on Industrial
  Electronics}, vol.~56, no.~7, pp. 2366--2373, 2009.

\bibitem{George2013}
L.~George and A.~Mazel, ``Humanoid robot indoor navigation based on 2d bar
  codes: Application to the nao robot,'' in \emph{Humanoid Robots (Humanoids),
  2013 13th IEEE-RAS International Conference on}.\hskip 1em plus 0.5em minus
  0.4em\relax IEEE, 2013, pp. 329--335.

\bibitem{engel2014lsd}
J.~Engel, T.~Sch{\"o}ps, and D.~Cremers, ``{LSD-SLAM: Large-scale direct
  monocular SLAM},'' in \emph{European Conference on Computer Vision}.\hskip
  1em plus 0.5em minus 0.4em\relax Springer, 2014, pp. 834--849.

\bibitem{mur2015orb}
R.~Mur-Artal, J.~M.~M. Montiel, and J.~D. Tardos, ``{ORB-SLAM: a versatile and
  accurate monocular SLAM system},'' \emph{IEEE Transactions on Robotics},
  vol.~31, no.~5, pp. 1147--1163, 2015.

\bibitem{endres20143}
F.~Endres, J.~Hess, J.~Sturm, D.~Cremers, and W.~Burgard, ``{3-D mapping with
  an RGB-D camera},'' \emph{IEEE Transactions on Robotics}, vol.~30, no.~1, pp.
  177--187, 2014.

\bibitem{misono2007development}
Y.~Misono, Y.~Goto, Y.~Tarutoko, K.~Kobayashi, and K.~Watanabe, ``{Development
  of laser rangefinder-based SLAM algorithm for mobile robot navigation},'' in
  \emph{SICE, 2007 Annual Conference}.\hskip 1em plus 0.5em minus 0.4em\relax
  IEEE, 2007, pp. 392--396.

\bibitem{klanvcar2014mobile}
G.~Klan{\v{c}}ar, L.~Tesli{\'c}, and I.~{\v{S}}krjanc, ``{Mobile-robot pose
  estimation and environment mapping using an extended Kalman filter},''
  \emph{International Journal of Systems Science}, vol.~45, no.~12, pp.
  2603--2618, 2014.

\bibitem{oliver2012using}
A.~Oliver, S.~Kang, B.~C. W{\"u}nsche, and B.~MacDonald, ``{Using the Kinect as
  a navigation sensor for mobile robotics},'' in \emph{Proceedings of the 27th
  conference on image and vision computing New Zealand}.\hskip 1em plus 0.5em
  minus 0.4em\relax ACM, 2012, pp. 509--514.

\bibitem{wang2016localization}
S.~Wang, Y.~Li, Y.~Sun, X.~Li, N.~Sun, X.~Zhang, and N.~Yu, ``{A localization
  and navigation method with ORB-SLAM for indoor service mobile robots},'' in
  \emph{Real-time Computing and Robotics (RCAR), IEEE International Conference
  on}.\hskip 1em plus 0.5em minus 0.4em\relax IEEE, 2016, pp. 443--447.

\bibitem{shiu1989calibration}
Y.~C. Shiu and S.~Ahmad, ``{Calibration of wrist-mounted robotic sensors by
  solving homogeneous transform equations of the form AX= XB},'' \emph{ieee
  Transactions on Robotics and Automation}, vol.~5, no.~1, pp. 16--29, 1989.

\bibitem{park1994robot}
F.~C. Park and B.~J. Martin, ``{Robot sensor calibration: solving AX= XB on the
  Euclidean group},'' \emph{IEEE Transactions on Robotics and Automation},
  vol.~10, no.~5, pp. 717--721, 1994.

\bibitem{fassi2005hand}
I.~Fassi and G.~Legnani, ``{Hand to sensor calibration: A geometrical
  interpretation of the matrix equation AX= XB},'' \emph{Journal of Field
  Robotics}, vol.~22, no.~9, pp. 497--506, 2005.

\bibitem{kelly2009fast}
J.~Kelly and G.~S. Sukhatme, ``{Fast relative pose calibration for visual and
  inertial sensors},'' in \emph{Experimental Robotics}.\hskip 1em plus 0.5em
  minus 0.4em\relax Springer, 2009, pp. 515--524.

\bibitem{hol2010modeling}
J.~D. Hol, T.~B. Sch{\"o}n, and F.~Gustafsson, ``{Modeling and calibration of
  inertial and vision sensors},'' \emph{The international journal of robotics
  research}, vol.~29, no. 2-3, pp. 231--244, 2010.

\bibitem{lu2014layered}
D.~V. Lu, D.~Hershberger, and W.~D. Smart, ``Layered costmaps for
  context-sensitive navigation,'' in \emph{Intelligent Robots and Systems (IROS
  2014), 2014 IEEE/RSJ International Conference on}.\hskip 1em plus 0.5em minus
  0.4em\relax IEEE, 2014, pp. 709--715.

\bibitem{fox1997dynamic}
D.~Fox, W.~Burgard, and S.~Thrun, ``The dynamic window approach to collision
  avoidance,'' \emph{IEEE Robotics \& Automation Magazine}, vol.~4, no.~1, pp.
  23--33, 1997.

\end{thebibliography}
}

\end{document}